\documentclass[10pt,journal,compsoc]{IEEEtran}
\usepackage{amsmath,bm}

\usepackage{amsfonts,amssymb}

\usepackage[noend]{algpseudocode}

\usepackage{algorithmicx,algorithm}

\usepackage{setspace}

\usepackage{multirow}

\usepackage{graphicx}

\ifCLASSOPTIONcompsoc
  \usepackage[nocompress]{cite}
\else
  \usepackage{cite}
\fi
\ifCLASSINFOpdf
\else
\fi
\begin{document}
%
\title{Two-Stage Copy-Move Forgery Detection with Self Deep Matching and Proposal SuperGlue}
%
%
%
%

\author{Yaqi~Liu,
        Chao~Xia,
        Xiaobin~Zhu,
        and~Shengwei~Xu
\IEEEcompsocitemizethanks{\IEEEcompsocthanksitem Y. Liu, C. Xia and S. Xu are with Beijing Electronic Science and Technology Institute, Beijing 100070, China.\protect\\
E-mail: liuyaqi@besti.edu.cn, xiachao@besti.edu.cn
\IEEEcompsocthanksitem X. Zhu is with the Department of Computer Science and Technology, School
of Computer and Communication Engineering, University of Science and
Technology Beijing, Beijing 100083, China.}
\thanks{(Corresponding author: Chao Xia.)}}
%
%
%

\markboth{}%
{Liu \MakeLowercase{\textit{et al.}}: Two-Stage Copy-Move Forgery Detection with Self Deep Matching and Proposal SuperGlue}
%



\IEEEtitleabstractindextext{%
\begin{abstract}
Copy-move forgery detection identifies a tampered image by detecting pasted and source regions in the same image. In this paper, we propose a novel two-stage framework specially for copy-move forgery detection. The first stage is a backbone self deep matching network, and the second stage is named as Proposal SuperGlue. In the first stage, atrous convolution and skip matching are incorporated to enrich spatial information and leverage hierarchical features. Spatial attention is built on self-correlation to reinforce the ability to find appearance similar regions. In the second stage, Proposal SuperGlue is proposed to remove false-alarmed regions and remedy incomplete regions. Specifically, a proposal selection strategy is designed to enclose highly suspected regions based on proposal generation and backbone score maps. Then, pairwise matching is conducted among candidate proposals by deep learning based keypoint extraction and matching, i.e., SuperPoint and SuperGlue. Integrated score map generation and refinement methods are designed to integrate results of both stages and obtain optimized results. Our two-stage framework unifies end-to-end deep matching and keypoint matching by obtaining highly suspected proposals, and opens a new gate for deep learning research in copy-move forgery detection. Experiments on publicly available datasets demonstrate the effectiveness of our two-stage framework. 
\end{abstract}

}

\maketitle

\IEEEdisplaynontitleabstractindextext

%
\IEEEpeerreviewmaketitle

\IEEEraisesectionheading{\section{Introduction}\label{sec:introduction}}

%
%
%
%
\IEEEPARstart{I}{ncreasing} availability and sophistication of digital image editing tools cause a major problem of that we even can not believe what we see \cite{liu2018image}. Image forgery is becoming a global epidemic which deeply affects our daily life for that some forgers use elaborately forged images to spread fake news or do other unscrupulous businesses \cite{wu2018busternet}. Copy-move forgery is a kind of image forgery in which one or several regions are pasted elsewhere in the same image in order to hide or duplicate objects of interest. Copy-move forgery detection techniques have always been a hot topic in image forensics \cite{ryu2010detection,christlein2012evaluation,li2015segmentation,cozzolino2015efficient}, and play important roles in cybersecurity and multimedia security \cite{qian2016separable,qiao2019adaptive}.

Conventional copy-move forgery detection methods adopt handcrafted features, and can be broadly divided into two categories, i.e., block-based approaches \cite{bashar2010exploring,ryu2013rotation,li2013image,li2013efficient,cozzolino2015efficient,mahmood2016copy,bi2018fast}, and keypoint-based approaches \cite{pan2010region,amerini2011sift,kakar2012exposing,pun2015image,li2015segmentation,ardizzone2015copy}. Their major difference is that block-based methods aim at exploring local features from abundant overlapping patches, while keypoint-based methods concentrate on patches of keypoints \cite{zandi2016iterative}. Nowadays, deep learning techniques have dominated various image processing tasks including image forensics \cite{cozzolino2017recasting,wu2017deep,liu2018image,cozzolino2018forensictransfer,zhou2018learning,cun2018image,cozzolino2019noiseprint,liu2019adversarial,yu2019attributing,mayer2019forensic,rossler2019faceforensics++}. Deep learning based copy-move forgery detection has also been investigated in \cite{wu2018image,wu2018busternet}. In \cite{wu2018image}, Wu et al. proposed an end-to-end deep neural network for predicting copy-move forgery masks. They construct a convolutional neural network for feature extraction, then compute self-correlation maps of convolutional features, and finally reconstruct forgery masks through a deconvolutional network. In \cite{wu2018busternet}, Wu et al. extended their network to a two-branch architecture: one branch localizes potential manipulation regions via visual inconsistencies; the other branch detects copy-move regions via visual similarities. According to the observations in \cite{korus2017multi,liu2018image,cozzolino2019noiseprint}, it is a very chanllenging task to localize forged regions in realistic forged images which barely have visual inconsistencies. As for the branch for detecting visual similarities, it tries to explore high-level low-resolution convolutional features \cite{simonyan2014very}, limiting the ability to detect accurate boundaries and small forged regions. Hence, we focus on digging deeper into visual similarity clues in our work.

\begin{figure*}[htp]
	\centering
	\includegraphics[width=15cm]{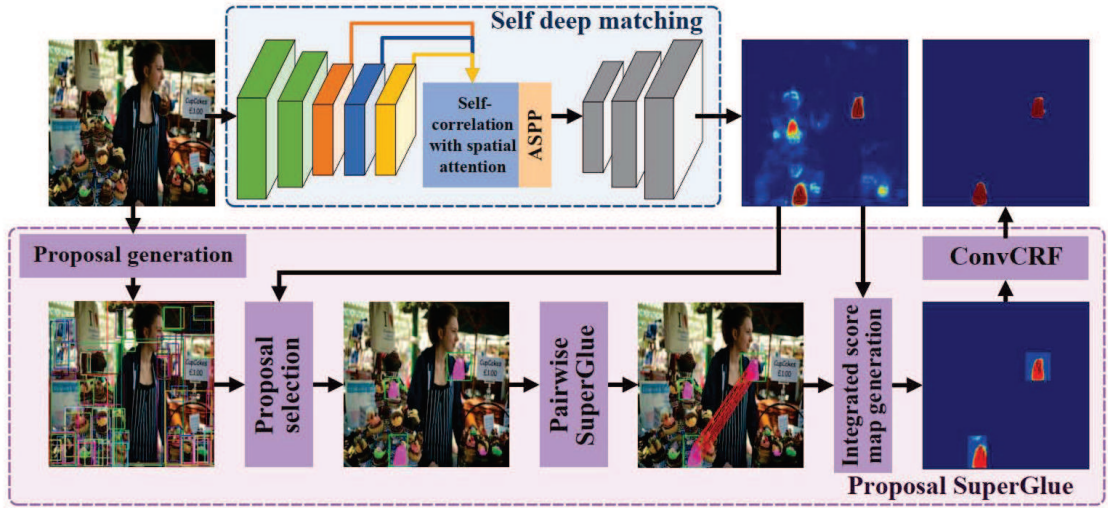}
	\caption{Overview of our two-stage copy-move forgery detection with self deep matching and Proposal SuperGlue. The first stage is a backbone self deep matching network based on atrous convolution, skip matching, and self-correlation with spatial attention. Red, blue, yellow blocks denote three convolutional blocks with the same scale, which are re-constructed by atrous convolution and skip matching. The second stage is named as Proposal SuperGlue. Proposal selection is conducted based on generated proposals and the backbone score map. Pairwise SuperGlue is conducted among candidate highly suspected proposals. Then, integrated score maps are generated by integrating matched keypoint scores and backbone scores. ConvCRF is constructed to refine integrated score maps.}
	\label{fig:framework}
\end{figure*}

In this paper, we propose a novel two-stage copy-move forgery detection framework which integrates end-to-end deep matching with proposal based keypoint matching. The pipeline of this two-stage framework is shown in Fig.~\ref{fig:framework}.

\textbf{In the first stage}, a backbone self deep matching network is constructed to generate backbone score maps which indicate suspicious probabilities of pixels. Our backbone network integrates atrous convolution, skip matching, and spatial attention. Atrous convolution can increase the resolution of feature maps, and skip matching can invesitgate hierarchical information. Particularly, we discover the inherent connections between spatial attention and self-correlation, and propose a self-correlation module with spatial attention. Previous copy-move forgery detection methods \cite{wu2018image,wu2018busternet} only adopt VGG16 \cite{simonyan2014very}, we further study the feasibility of constructing deep matching based on deeper networks (ResNet50, ResNet101 \cite{he2016deep}) and light-weight networks (MobileNet \cite{howard2017mobilenets,sandler2018mobilenetv2,howard2019searching}, ShuffileNet \cite{zhang2018shufflenet,ma2018shufflenet}). The backbone network is regarded as a filter to efficiently detect suspected forged regions, while the results may inevitably contain false-alarmed regions or incomplete regions. Thus, we propose the second stage to remove false-alarmed regions and remedy incomplete regions.

\textbf{In the second stage}, a proposal based keypoint matching method is proposed and named as Proposal SuperGlue. Proposal SuperGlue mainly consists of two components: (1) A proposal selection module obtains several highly suspected proposals from a large number of bounding boxes provided by a proposal generation method \cite{pinheiro2015learning}. Proposal selection takes advantage of both backbone score maps and appearance clues, bridging the gap between deep matching and keypoint matching. (2) Proposal matching and label generation are devised to remove false alarms, remedy incomplete regions, and generate pixel labels from score maps. Deep-learning keypoint extraction (SuperPoint \cite{detone2018superpoint}) and matching (SuperGlue \cite{sarlin2020superglue}) are conducted among candidate proposals. An integrated score map generation method is designed to integrate keypoint matching results and backbone score maps. And an integrated score map refinement method is presented based on an improved fully connected CRF, i.e., ConvCRF (Convolutional Conditional Random Field) \cite{teichmann2018convolutional}.

Specifically, our main contributions of
this paper can be summarized as follows:
\begin{itemize}
	\item An innovative two-stage copy-move forgery detection framework is proposed based on self deep matching and Proposal SuperGlue. We imaginatively integrate end-to-end deep matching with keypoint matching through highly suspected proposals.
	\item A backbone self deep matching network is constructed based on atrous convolution, skip matching and spatial attention. Inherent connections between self-correlation and spatial attention are elaborately investigated.
	\item Proposal SuperGlue, which incorporates proposal generation and deep-learning keypoint matching with a series of postprocessing procedures, is proposed to effectively remove false-alarmed regions and remedy incomplete regions. 
\end{itemize}

The structure of this paper is as follows: In Section \ref{sec:rw},
we discuss related work. In Section \ref{sec:methodology}, we elaborate the
proposed framework. In Section \ref{sec:experiment}, experiments are conducted. In Section \ref{sec:conclusion}, we draw conclusions.

\section{Related Work}
\label{sec:rw}

In this section, we briefly review the state-of-the-art copy-move forgery detection methods, attention mechanism, proposal generation and local feature matching which are the key techniques researched in our work. 

\textbf{Copy-move forgery detection}. Conventional copy-move forgery detection methods mainly consist of three components \cite{cozzolino2015efficient}: (1) feature extraction: extracting suitable features from pixels of interest; (2) matching: computing their best matching based on their associated features; (3) post-processing: processing and filtering vague detections to reduce false alarms. According to the formulations of feature extraction and subsequent matching schemes, these methods can be classified into two categories, i.e., block-based and keypoint-based methods. In block-based methods, a variety of features have been investigated for describing overlapping blocks and dense matching, e.g., DCT (Discrete Cosine Transform) \cite{mahmood2016copy}, DWT (Discrete Wavelet Transform) and KPCA (Kernel Principal Component Analysis) \cite{bashar2010exploring},
Zernike moments \cite{ryu2013rotation}, PCT (Polar Cosine Transform) \cite{yap2010two,li2013image}, PCET (Polar Complex Exponential Transform) \cite{bi2018fast}, LBP (Local Binary Patterns) \cite{li2013efficient}, Circular Harmonic Transforms (CHT) \cite{cozzolino2015efficient}.  In keypoint-based methods, the commonly used features are SIFT (Scale Invariant Feature Transform) \cite{pan2010region,amerini2011sift,li2015segmentation,pun2015image} and SURF (Speeded-Up Robust Features) \cite{ardizzone2015copy,silva2015going}. Although great progress has been made in the study of copy-move forgery detection, it is still an unresolved challenging task for that duplicate regions may be small or smooth,  and have gone through complicated rotation, resizing, compression and noise addition \cite{bi2018fast}. Besides, all the above conventional copy-move forgery detection methods rely on hand-crafted features and each module is optimized independently \cite{wu2018busternet}. Consequently, two kinds of end-to-end deep learning based copy-move forgery detection methods were proposed by Wu et al. in \cite{wu2018image,wu2018busternet}.

\textbf{Attention mechanism}. In \cite{sutskever2014sequence}, Sutskever et al. constructed a multi-layer long short term memory (LSTM) to map the input sequence to a fixed-length vector, and another deep LSTM to decode the target sequence from the vector. In \cite{bahdanau2014neural}, Bahdanau et al. adopted the attention mechanism to dynamically generate the vectors. Since then, the attention mechanism has been widely applied to solve sequential decision tasks \cite{lin2017structured}, and numerous attention-based models have been proposed \cite{xu2015show,luong2015effective,vaswani2017attention}. The attention mechanism can bias the allocation of available processing resources to the most informative components of input signals \cite{hu2018squeeze}, and has also been applied to solve multimedia problems, e.g., image classification \cite{hu2018squeeze,wang2017residual}, object detection \cite{woo2018cbam}, image super-resolution \cite{zhang2018image}, video classification \cite{wang2018non}. In these tasks, consistent improvements have been gained by adopting attention mechanisms to recalibrate informative convolutional features. 

\textbf{Proposal generation}. In our work, we try to generate bounding boxes enclosing suspected regions. Proposal generation is a kind of technique that has been widely researched before the arrival of end-to-end object detection \cite{ren2015faster}. It aims to find out a set of (ranging from hundreds to thousands per image) proposal regions or bounding boxes which may contain objects \cite{liu2017listnet}. Since we can get hundreds of proposals which are near to contours in images, they cover suspected regions with high probability. Proposal generation approaches can be divided into two categories: conventional methods and deep-learning methods. Conventional methods leverage low-level grouping and saliency clues, e.g., objectness scoring \cite{alexe2012measuring,zitnick2014edge}, seed segmentation \cite{humayun2014rigor,krahenbuhl2014geodesic,krahenbuhl2015learning}, superpixel merging \cite{uijlings2013selective,pont2016multiscale}. Deep-learning approaches construct deep-network architectures to obtain proposals. For example, Deepbox \cite{kuo2015deepbox} learns a convolutional network to rerank proposals generated by EdgeBox \cite{zitnick2014edge}. Multibox \cite{erhan2014scalable} constructs a deep network to generate bounding box proposals. DeepMask \cite{pinheiro2015learning} and SharpMask \cite{pinheiro2016learning} can generate and refine segmentation proposals with high efficiency. These deep-learning approaches give us an opportunity to efficiently generate suspected boxes enclosing forged regions from a small set of candidate proposals.

\textbf{Local feature matching}. It mainly consists of five steps, (1) detecting interest points, (2) computing visual descriptors, (3) matching visual descriptors with a nearest neighbor (NN) search, (4) filtering incorrect matches, (5) estimating a geometric transformation \cite{sarlin2020superglue}. In recent years, researchers have been trying to learn better sparse detectors and local descriptors \cite{detone2018superpoint,dusmanu2019d2,DBLP:conf/nips/RevaudSHW19,ono2018lf,yi2016lift} from data using Convolutional Neural Networks (CNNs), and attempting to improve their discriminative ability by using various strategies, e.g., a wider context using regional features, log-polar patches, unsupervised learning. Although tremendous progress has been made in this field, these sets of matches are still estimated by NN search. In \cite{sarlin2020superglue}, a novel approach based on graph neural networks, i.e., SuperGlue, is proposed to establish pointwise correspondences from off-the-shelf local features: it acts as a middle-end between hand-crafted or learned front-end and back-end. SuperGlue outperforms other learned approaches and achieves state-of-the-art results on pose estimation. In keypoint-based copy-move forgery detection approaches \cite{pan2010region,amerini2011sift,kakar2012exposing,pun2015image,li2015segmentation,ardizzone2015copy}, hand-crafted local features and NN search have been widely researched. In our work, we try to integrate learning based detector, discriptor and matching into a unified copy-move forgery detection framework.


\section{Methodology}
\label{sec:methodology}

Our two-stage copy-move forgery detection framework consists of self deep matching and Proposal SuperGlue, as shown in Fig. \ref{fig:framework}. The first stage is a backbone self deep matching network, which generates score maps in an end-to-end manner. Firstly, we introduce the main architecture of our backbone network, including several alternative formulations in section \ref{sssec:FEAC}. Then, we introduce self-correlation with spatial attention in section \ref{sssec:SCSA}. The second stage is called Proposal SuperGlue, which is proposed to remove false-alarmed regions and remedy incomplete regions. In section \ref{sssec:psel}, we introduce our proposal selection strategy based on deep-learning proposal generation. In section \ref{sssec:PMLG}, we introduce proposal-based point matching, integrated score map generation and refinement.

\subsection{Self Deep Matching}
\label{ssec:sdmarch}

\subsubsection{Backbone Network Architecture}
\label{sssec:FEAC}

\begin{figure}[htbp]
	\centering
	\includegraphics[width=0.86\columnwidth]{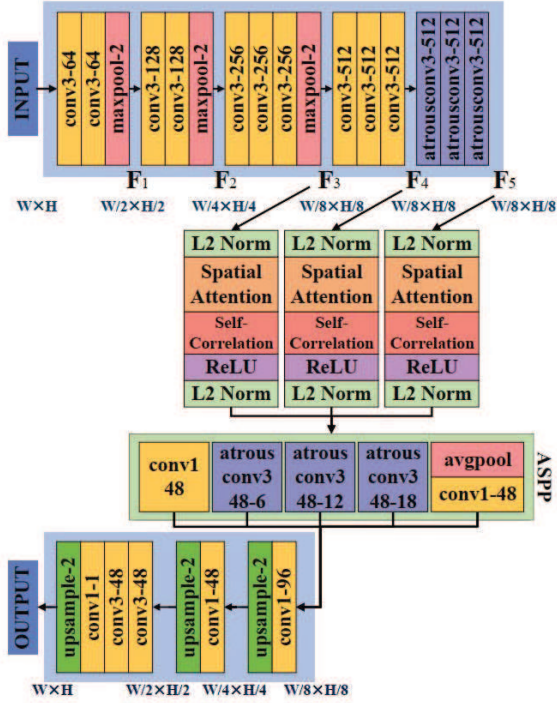}
	\caption{The architecture of backbone network with VGG16.}
	\label{Figure:ADMarch}
\end{figure}

In our work, we adopt VGG16 as our basic feature extractor, remove pooling operatons in the fourth and fifth convolutional blocks \cite{simonyan2014very}, and adjust the fifth block by adopting atrous convolution to keep their original field-of-views. Atrous convolution can generalize standard convolution, adjust filter’s field-of-view and control the resolution of convolutional features \cite{chen2018deeplab,chen2018encoder,chen2017rethinking}. Let $\mathbf{y}(i_c,j_c)$ denote the output of the atrous convolution of a $2$-D input signal $\mathbf{x}(i_c,j_c)$, and the atrous convolution can be computed as:
\begin {equation}\label{eq:yij}
\mathbf{y}(i_c,j_c)=\sum_{k_1,k_2}\mathbf{w}(k_1,k_2)\times \mathbf{x}(i_c+r_\mathrm{ac} k_1,j_c+r_\mathrm{ac} k_2)
\end {equation}
where $k_1,k_2\in[-fl(\frac{K}{2}),fl(\frac{K}{2})]$ ($fl(\cdot)$ is a floor function), $\mathbf{w}(k_1,k_2)$ denotes a $K\times K$ filter, atrous rate $r_\mathrm{ac}$ determines the stride with which we sample the input
signal. In the fifth block of our basic architecture, atrous rate $r_\mathrm{ac}$ is set to $2$. Consequently, we generate three groups of larger feature maps with the same size, i.e., $\mathbf{F}_3$, $\mathbf{F}_4$ and $\mathbf{F}_5$ in Fig. \ref{Figure:ADMarch}. These hierarchical feature maps are all fed into our self-correlation module with spatial attention to compute the correlation maps. This kind of skip connections between multi-level feature maps and correlation computation is named as skip matching, and can effectively leverage rich hierarchical information provided by the feature extractor. And in the next section, we will introduce our self-correlation with spatial attention in detail.

Based on the computed correlation maps, we construct an Atrous Spatial
Pyramid Pooling (ASPP) module to capture their multiscale information. As shown in Fig.~\ref{Figure:ADMarch}, we construct $3$ parallel atrous convolutional layers with $3\times 3$ filters and atrous rates of $\{6,12,18\}$. Besides, we construct a convolutional layer with $1\times 1$ filters, and a global average pooling layer followed by a convolutional layer with $1\times 1$ filters to capture local features and image-level features respectively. All these convotional layers output $48$-channel feature maps, and the five groups of feature maps are concatenated and fed into subsequent layers which are constituted of convolutional and upsampling layers.

\textbf{Alternative formulations}. Previous end-to-end copy-move forgery detection approaches \cite{wu2018image,wu2018busternet} adopt VGG16 for feature extraction, and we do not know the performance of deeper or light-weight networks. Thus, we formulate the popular ResNet50 and ResNet101 \cite{he2016deep} as deeper feature extractor, decrease the strides of fourth and fifth convolutional blocks, and set the atrous rates as $2$ and $4$ respectively. Light-weight networks are specifically tailored for mobile and resource constrained environments \cite{sandler2018mobilenetv2}. We reformulate three popular and competitive light-weight networks, i.e., MobileNetV2 \cite{sandler2018mobilenetv2}, MobileNetV3 \cite{howard2019searching} and ShuffleNetV2 \cite{ma2018shufflenet}. We enlarge feature maps of the last two convolutional blocks by decreasing strides and adopting atrous convolution. In these formulations, we still can get $3$ sets of feature maps with the same size. And these features are fed into correlation layers and subsequent score map generation layers.

\subsubsection{Self-Correlation with Spatial Attention}
\label{sssec:SCSA}

In this section, we detailedly introduce the proposed self-correlation with spatial attention, and discuss the conections between self-correlation and spatial attention. Let $\mathbf{F}_{l}$ denote the $l$-th block feature maps, and $\mathbf{F}_{l}(i,j)$ denotes a $c$-dimensional descriptor at $(i,j)$. Note that $\mathbf{F}_{l}\in\mathbb{R}^{h\times w \times c}$, $i\in[1,h]$, $j\in[1,w]$, $h$ and $w$ indicate the height and width of the feature map, and $h=w$ in our work. Before the attention and correlation computation, L2-normalization is conducted, $\bar{\mathbf{F}}_{l}(i,j)=\mathrm{L2\_norm}(\mathbf{F}_{l}(i,j))={\mathbf{F}_{l}(i,j)}/{||\mathbf{F}_{l}(i,j)||_2}$. By adopting L2-normalization, we can restrict the value ranges of descriptors, and obtain normalized feature maps, i.e., $\bar{\mathbf{F}}_{l}$.

Spatial attention is a kind of self-attention module which calculates response at a position as a weighted sum of the features at all positions \cite{vaswani2017attention}. It can capture long-range dependences and allocate attention according to
similarity of color and texture. In our work, spatial attention is constructed to reinforce $\bar{\mathbf{F}}_{l}$. $\bar{\mathbf{F}}_{l}$ is first transformed into two feature spaces $\bm{f}(\bar{\mathbf{F}}_{l})=\bar{\mathbf{F}}_{l}\bm{W}_{l,\bm{f}}+\bm{b}_{l,\bm{f}}$ and $\bm{g}(\bar{\mathbf{F}}_{l})=\bar{\mathbf{F}}_{l}\bm{W}_{l,\bm{g}}+\bm{b}_{l,\bm{g}}$. Then we compute:
\begin {equation}\label{eq:spatialbeta}
\beta_l^{(m,n)}=\frac{\mathrm{exp}(s_l^{(m,n)})}{\sum_n\mathrm{exp}(s_l^{(m,n)})}
\end {equation}
where 
\begin {equation}\label{eq:spatialbetasmn}
s_l^{(m,n)}= \bm{f}(\bar{\mathbf{F}}^{(m)}_{l})^T\bm{g}(\bar{\mathbf{F}}^{(n)}_{l})
\end {equation}
$\beta_l^{(m,n)}$ indicates the extent to which the model attends to the $n$-th location when predicting the $m$-th region, $m,n\in[1,h\times w]$. The output of the attention block is computed as:
\begin {equation}\label{eq:spatialatten}
\mathbf{o}_l^{(m)}=\sum_n\beta_{mn}\bm{h}(\bar{\mathbf{F}}^{(n)}_{l})
\end {equation}
where $\bm{h}(\bar{\mathbf{F}}_{l})=\bar{\mathbf{F}}_{l}\bm{W}_{l,\bm{h}}+\bm{b}_{l,\bm{h}}$. Note that $\bm{W}_{l,\bm{f}}\in\mathbb{R}^{c \times \frac{c}{8}}$, $\bm{W}_{l,\bm{g}}\in\mathbb{R}^{c \times \frac{c}{8}}$, $\bm{W}_{l,\bm{h}}\in\mathbb{R}^{c \times c}$, $\bm{b}_{l,\bm{f}}\in \mathbb{R}^{\frac{c}{8}}$, $\bm{b}_{g,\bm{h}}\in \mathbb{R}^{\frac{c}{8}}$, $\bm{b}_{l,\bm{h}}\in \mathbb{R}^{c}$, which are implemented as $1\times 1$ convolutional layers. $\mathbf{O}_l=\{\mathbf{o}_l^{(1)},\mathbf{o}_l^{(2)},\cdots,\mathbf{o}_l^{(h\times w)}\}$ are the attention values for $\bar{\mathbf{F}}_l$. Consequently, the spatial attention reinforced convolutional feature maps can be computed as:
\begin {equation}\label{eq:sattwefea}
\ddot{\mathbf{F}}_l=\mathrm{Atten}_l(\bar{\mathbf{F}}_l)=\lambda_l\mathbf{O}_l+\bar{\mathbf{F}}_l
\end {equation}
where $\lambda_l$ is a scale parameter which is initialized as $0$ and gradually learned to assign a proper value.

Self-correlation aims to compute the similarity between every two locations in the convolutional feature maps. Scalar product is commonly used:
\begin {equation}\label{eq:selfcorr}
c_l^{(m,n)}=(\ddot{\mathbf{F}}^{(m)}_{l})^T \ddot{\mathbf{F}}^{(n)}_{l}
\end {equation}
Thus, we can get a raw correlation map tensor $\mathbf{C}_l=\{c_l^{(m,n)}|m,n\in[1,h\times w]\}\in \mathbb{R}^{h\times w \times (h\times w)}$. In fact, only a small fraction of features has close relations, and the majority of features are dissimilar. This indicates that a subset of $\mathbf{C}_l$ contains sufficient information to decide which feature is matched. Consequently, $\mathbf{C}_l$ is sorted along the $(h\times w)$ channels, and top-$T$ values are selected:
\begin {equation}\label{eq:sortpool}
\tilde{\mathbf{C}}_l(i,j,1:T)=\mathrm{Top\_T}(\mathrm{Sort}(\mathbf{C}_l(i,j,:)))
\end {equation}
A monotonic decreasing curve with an abrupt drop at some point should be observed along the $T$ channels, as long as $\tilde{\mathbf{C}}_l(i,j)$ has matched regions. Thus, the $T$ channels should cover the most drops, and the selection of $T$ is discussed in experiments. In theory, our network can process arbitrary-sized images since the adoption of the top-$T$ selection, unless $h\times w<T$.  Moreover, zero-out and normalization operations are conducted on $\tilde{\mathbf{C}}_l$ to limit correlation values to certain ranges and filter redundant values:
\begin {equation}\label{eq:relunorm}
\bar{\mathbf{C}}_l=\mathrm{L2\_norm}(\mathrm{Max}(\tilde{\mathbf{C}}_l,0))
\end {equation}

Since we get three groups of feature maps with the same size from feature extractor, i.e., $l\in\{3,4,5\}$, we can get three groups of correlation maps, i.e., $\bar{\mathbf{C}}_3$, $\bar{\mathbf{C}}_4$ and $\bar{\mathbf{C}}_5$. Note that the parameters of spatial attention are not shared for the computation of these three groups of correlation maps. Then, we concatenate the three groups of correlation maps, and get a correlation map tensor $\widehat{\mathbf{C}}=\mathrm{Concat}(\bar{\mathbf{C}}_3,\bar{\mathbf{C}}_4,\bar{\mathbf{C}}_5)$, where $\widehat{\mathbf{C}}\in\mathbb{R}^{h \times w \times 3T}$. Since $\widehat{\mathbf{C}}$ is computed from three groups of hierarchical feature maps, it contains rich correlation relations from coarse to fine.

\textbf{Inherent connections between self-correlation and spatial attention.} Both self-correlation and spatial attention attempt to explore correlations between every pair of features in a feature map tensor. Spatial attention conducts a scalar product in transformed spaces as Eq. (\ref{eq:spatialbetasmn}), while self-correlation conducts a scalar product directly on feature maps as Eq. (\ref{eq:selfcorr}). Essentially, they have the same target of finding the close related regions. Spatial attention can allocate attention according to similarity of features, driving correlated regions to have closer feature distributions. Inspired by this inherent connection, we construct spatial attention before self-correlation computation. Consequently, it can reinforce the subsequent self-correlation computation. 

Additionally, we have also investigated multi-head spatial attention. Multi-head attention allows the model to jointly attend to information from different representation subspaces at different positions \cite{vaswani2017attention}. In fact, it constructs several parallel spatial attention blocks. Furthermore, we also attempt to add channel attention (SE blocks \cite{hu2018squeeze}) before or after spatial attention. Channel attention can highlight channel-wise informative
features, and a weighted scalar
product can be conducted in the correlation computation
procedure. However, they can not achieve better performance with additional parameters which will be discussed in experiments.

\subsection{Proposal SuperGlue}
\label{ssec:psg}

Proposal SuperGlue can be broadly divided into two steps. \textit{In the first step}, a proposal selection method is proposed to obtain highly suspected regions from hundreds of bounding-box proposals. These bounding-box proposals are generated by a proposal generation method which exploits image appearance features to find bounding boxes near to contours. \textit{In the second step}, we devise proposal-based keypoint matching with elaborately designed postprocessing procedures. Firstly, pairwise deep-learning keypoint matching is conducted among candidate proposals. Then, we propose an integrated score map generation method to integrate both self deep matching and keypoint matching results, so that some false-alarmed regions can be removed and incomplete regions can be complemented. Finally, in order to get good score distributions and accurate boundaries, ConvCRF is constructed to refine integrated score maps according to integrated scores and associated apperance similarity in the image. \textit{The first step} is introduced in section \ref{sssec:psel}, and \textit{the second step} is discussed in section \ref{sssec:PMLG}.

\subsubsection{Proposal Selection}
\label{sssec:psel}

Processed by our backbone self deep matching network, we can get a score map $\mathbf{S}\in\mathbb{R}^{h_I\times w_I}$, $h_I$ and $w_I$ denote the size of the score map which is the same as the size of input image $\mathbf{I}$. $\mathbf{S}(i_I,j_I)\in[0,1]$, and $i_I\in [1,h_I],j_I\in [1,w_I]$. Since the feature maps of the backbone network have lower resolutions than original images, and the self-correlation is built on primitive scalar product instead of intricate similarity computing, $\mathbf{S}$ may have false-alarmed or incomplete regions. In order to get rid of false-alarmed regions and complement incomplete regions, a proposal selection strategy is proposed to obtain highly suspected boxes for further matching.

Our motivation is that $\mathbf{S}$ always has some isolated meaningless regions in some complicated images according to our observation. Whether we can enclose meaningful regions while ignore meaningless regions may affect the performance. However, how can we obtain several well enclosed bounding boxes based on score map $\mathbf{S}$ and input image $\mathbf{I}$? After all, there are too many bounding boxes can be generated from a single image, e.g., different scales, aspect ratios and positions. In fact, the majority of copy-move forged regions have clear contours, and might be possible to be covered by genereted proposals which are relied on edges or saliency features \cite{liu2017listnet}. Thus, we propose to conduct proposal generation on the input image, and select several high-quality boxes from hundreds of proposals. Our proposal selection strategy relies on score map $\mathbf{S}$, and consists of selecting and merging operations. We assume that there is a proposal generation function $\mathcal{P}(\cdot)$ with image $\mathbf{I}$ as input, we can get $P$ proposals $\mathbf{P}=\{\bm{p}_p|p\in[1,P]\}$. $\bm{p}_p=\{(x^p_1,y^p_1),(x^p_2,x^p_2)\}$ contains the coordinates of top left and bottom right corners. With score map $\mathbf{S}$ at hand, we can get the average score in proposal $\bm{p}_p$, i.e., $s_p = \mathnormal{f}_{\mathrm{avgs}}(\mathbf{S},\bm{p}_p)$. According to $s_p$ and its relations with other proposals, we can obtain the final proposals. Our proposal selection strategy can be summarized as Algorithm~\ref{pssalg}.

\begin{algorithm}[htp]
	\caption{Proposal selection strategy.}
	\label{pssalg}
	\hspace*{0.02in} {\bf Input:} Image
	 $\mathbf{I}$ and score map $\mathbf{S}$\\
	\hspace*{0.02in} {\bf Output:} Selected proposals $\mathbf{P}_{s}$
	\begin{algorithmic}[1]
		\State $\mathbf{P}=\mathcal{P}(\mathbf{I})$;
		\State $\mathbf{P}_{t}=\{\}$;
		\For{$p=1$ to $P$}		
		\State $s_p = \mathnormal{f}_{\mathrm{avgs}}(\mathbf{S},\bm{p}_p)$;
		\If{$s_p > s_{t}$}
		\State $flag=1$;
		\For{$i_{ps}=1$ to $\mathrm{len}(\mathbf{P}_{t})$}
		\State $v_{iou}=\mathrm{IoU}(\mathbf{P}_{t}(i_{ps}),\bm{p}_p)$;
		\State $v_{inter}=\mathrm{Inter}(\mathbf{P}_{t}(i_{ps}),\bm{p}_p)$;
		\If{$v_{iou}>iou_{t}$} 
		\If {$s_p>\mathnormal{f}_{\mathrm{avgs}}(\mathbf{S},\mathbf{P}_{t}(i_{ps}))$}
		\State $\mathbf{P}_{t}(i_{ps})=\bm{p}_p$; $flag=0$; Break;
		\EndIf
		\EndIf
		\If {$v_{inter}/\mathrm{Size}(\bm{p}_p)>inter_t$ or \\\hspace*{0.8in}$v_{inter}/\mathrm{Size}(\mathbf{P}_{t}(i_{ps}))>inter_t$}
		\State $\bm{p}_{m}=\mathrm{Merge}(\bm{p}_p,\mathbf{P}_{t}(i_{ps}))$;
		\If {$\mathnormal{f}_{\mathrm{avgs}}(\mathbf{S},\bm{p}_{m})>s_{t}$}
		\State $\mathbf{P}_{t}(i_{ps})=\bm{p}_{m}$; $flag=0$; Break;
		\EndIf
		\EndIf
		\EndFor
		\If{$flag=1$}
		\State $\mathbf{P}_{t}=\mathbf{P}_{t}\cup\bm{p}_p$;
		\EndIf
		\EndIf   		
		\EndFor
		\State $\mathbf{P}_{s}=\{\}$;
		\While{$\mathrm{len}(\mathbf{P}_{s})\ne \mathrm{len}(\mathbf{P}_{t})$}
		\If {$\mathbf{P}_{s}\ne\oslash$}
		\State $\mathbf{P}_{t}=\mathbf{P}_{s}$;
		\EndIf
		\State $\mathbf{P}_{s}=\{\}$;
		\For{$i_{ps1}=1$ to $\mathrm{len}(\mathbf{P}_{t})$}
		\For{$i_{ps2}=i_{ps1}+1$ to $\mathrm{len}(\mathbf{P}_{t})$}
		\State $v_{inter}=\mathrm{Inter}(\mathbf{P}_{t}(i_{ps1}),\mathbf{P}_{t}(i_{ps2}))$;
		\If {$v_{inter}/\mathrm{Size}(\mathbf{P}_{t}(i_{ps1}))>inter_t$ or\\\hspace*{0.6in} $v_{inter}/\mathrm{Size}(\mathbf{P}_{t}(i_{ps2}))>inter_t$}
		\State $\bm{p}_{m}=\mathrm{Merge}(\mathbf{P}_{t}(i_{ps1}),\mathbf{P}_{t}(i_{ps2}))$;
		\If {$\mathnormal{f}_{\mathrm{avgs}}(\mathbf{S},\bm{p}_{m})>s_{t}$}
		\State $\mathbf{P}_{s}=\mathbf{P}_{s}\cup\bm{p}_{m}$;
		\Else
		\State Insert $\mathbf{P}_{t}(i_{ps1})$ or $\mathbf{P}_{t}(i_{ps2})$ \\\hspace*{1.0in}with higher intersection rate;
		\EndIf
		\EndIf
		\EndFor
		\EndFor
		\EndWhile
	\end{algorithmic}
\end{algorithm}

In Algorithm~\ref{pssalg}, there are some basic functions: $\mathrm{len}(\cdot)$ returns the item number of input list, $\mathrm{IoU}(\cdot,\cdot)$ computes the Intersection over Union (IoU) \cite{liu2017listnet} of two boxes, $\mathrm{Inter}(\cdot,\cdot)$ computes their intersection, and $\mathrm{Size}(\cdot)$ indicates the size of the input box. $\mathrm{Merge}(\cdot,\cdot)$ is used to merge two input boxes, in other words, it generates the smallest box which can cover the two input boxes. The proposal generation function $\mathcal{P}(\cdot)$ is implemented based on DeepMask \cite{pinheiro2015learning}. The basic idea of Algorithm~\ref{pssalg} is that we try to reject proposals with small average scores, select proposals with higher scores from proposals which have high IoU with each other, and merge proposals or select larger boxes when they have large intersection rates. And there are some parameters need to set, proposal threshold score $s_{t}=0.4$, threshold IoU $iou_t=0.5$, threshold intersection rate $inter_t=0.8$. Besides, all proposals or merged boxes should meet the basic requirement that they should be smaller than the half of the input image. Last but not least, iterations are conducted to avoid merged boxes with large intersection rates. By using Algorithm~\ref{pssalg}, we can get $\tilde{P}$ (generally less than $10$) high-quality proposals $\mathbf{P}_{s}=\{\bm{p}_{\tilde{p}}|\tilde{p}\in[1,\tilde{P}]\}$ ($\mathbf{P}_{s}(\tilde{p})$ indicates $\bm{p}_{\tilde{p}}$ in $\mathbf{P}_{s}$).

\subsubsection{Keypoint Matching and Label Generation}
\label{sssec:PMLG}

\textbf{Proposal-based keypoint matching}. With high-qualtiy proposals $\mathbf{P}_{s}$ at hand, we can extract interest points from them and conduct keypoint matching. As we discussed in Section \ref{sec:rw}, CNN-based interest point detection and discription show a good prospect in numerous applications. Thus, we extract keypoints with corresponding descriptors from each proposal using SuperPoint \cite{detone2018superpoint}. It is a fully-convolutional model which operates on full-sized
images and jointly computes pixel-level interest point locations and associated descriptors in one forward pass. It can be denoted as $\mathbf{K}_{\tilde{p}}=\mathrm{SuperPoint}(\bm{p}_{\tilde{p}},\mathbf{I})$, where $\mathbf{K}_{\tilde{p}}$ denotes the extracted keypoints set from proposal $\bm{p}_{\tilde{p}}$, and the $k_p$-th elements is $\mathbf{K}_{\tilde{p}}(k_p)=\{(x_{k_p},y_{k_p}),\mathbf{d}_{k_p}\}$, $\mathbf{d}_{k_p}$ denotes the corresponding descriptor. Then for each pair of point sets $\mathbf{K}_{\tilde{p}}(k_{p1})$ and $\mathbf{K}_{\tilde{p}}(k_{p2})$, we conduct SuperGlue \cite{sarlin2020superglue} to get matched points and corresponding matching scores $\mathbf{M}_{\tilde{p}1},\mathbf{M}_{\tilde{p}2}=\mathrm{SuperGlue}(\mathbf{K}_{\tilde{p}1},\mathbf{K}_{\tilde{p}2})$. SuperGlue uses a graph neural network and attention to solve an assignment optimization problem. Instead of learning better task-agnostic local features followed by simple matching heuristics and tricks, SuperGlue learns the matching process from pre-existing local features using a novel neural architecture for the first time. It matches two sets of local features by jointly finding correspondences and rejecting non-matchable points. Finally, we can get $M$ matched points and their matching scores $\mathbf{M}=\{\mathbf{M}_{\tilde{p}}|\tilde{p}\in[1,\tilde{P}]\}=\{(x_m,y_m),s_m|m\in[1,M]\}$.

\textbf{Integrated score map generation}. In order to map the matching scores to each pixel, we adopt a superpixel algorithm, i.e., SEEDS \cite{van2012seeds,van2015seeds}. By conducting superpixel segmentation, we get the superpixel labels $\mathbf{L}_{sp} = \mathrm{SuperPixel}(\mathbf{I})$. Let  $\mathbf{L}_{sp}(x_m,y_m)$ denote pixels whose superpixel labels are the same as $(x_m,y_m)$. We set scores of these pixels the same as the score of matched point $(x_m,y_m)$, i.e., $\mathbf{S}_{sp}(\mathbf{L}_{sp}(x_m,y_m))=s_m$. Thus, we get our pixel-level scores $\mathbf{S}_{sp}$ from superpixel and matched points. Besides, we also generate a pixel-level score map $\mathbf{S}_{p}$ from backbone scores $\mathbf{S}$ and candidate proposals which have matched points. Concretely, we set $\mathbf{S}_{p}(x,y)=\mathbf{S}(x,y)$ for $(x,y)$ in the scope of $\bm{p}_{\tilde{p}}$ which contains matched points, otherwise $\mathbf{S}_{p}(x,y)=0$. Thus, our integrated score map $\mathbf{S}_{in}$ is computed based on $\mathbf{S}_{sp}$ and $\mathbf{S}_{p}$ as follows:
\begin {equation}\label{eq:finalscore}
\mathbf{S}_{in}=\frac{1}{1+\mathrm{exp}(-\phi(\alpha\cdot\mathbf{S}_{sp}+\beta\cdot\mathbf{S}_{p}+\gamma))}
\end {equation}
where $\alpha$, $\beta$ and $\gamma$ are three parameters to balence $\mathbf{S}_{sp}$ and $\mathbf{S}_{p}$. We set $\alpha=\beta=1$ to make $\mathbf{S}_{sp}$ and $\mathbf{S}_{p}$ have the same contribution. We set $\gamma=-0.5$ to make sure it has the same distribution when $\mathbf{S}_{sp}(i)=0$. $\phi$ indicates the amplifying factor to control score distribution of $\mathbf{S}_{in}$, and is set to $4$.

\textbf{Integrated score map refinement for label generation}. The directly computed $\mathbf{S}_{in}$ has some small isolated regions or holes inside detected regions, because there are some false-alarmed or missing-detected regions. In order to neglect regions with lower matching probability and refine contours according to image content, we formulate fully connected CRF (Conditional Random Field) \cite{krahenbuhl2011efficient} based on $\mathbf{S}_{in}$ and image $\mathbf{I}$, to get final labels. Our problem is that we have an image $\mathbf{I}$ which has $N$ pixels, and we try to fulfill a segmentation task with two classes. A segmentation of $\mathbf{I}$ is modelled as a random field $\mathbf{X}=\{X_1,\cdots,X_N\}$ where each random variable $X_n$ takes values of $\{0,1\}$. ``$1$" is used to label forged locations and corresponding genuine ones, while ``$0$" is for remaining parts. A conditional random field $(\mathbf{I}, \mathbf{X})$ is characterized by a Gibbs distribution $P(\mathbf{X}|\mathbf{I})=\frac{1}{Z(\mathbf{I})}\mathrm{exp}(-E(\mathbf{X}|\mathbf{I}))$, where the energy function $E(\mathbf{X}|\mathbf{I})$ is given by:
\begin {equation}\label{eq:crfenergy}
E(\mathbf{X}|\mathbf{I})=\sum_{i\le N}{\psi_u(X_i|\mathbf{I})} + \sum_{i\ne j\le N}{\psi_p(X_i,X_j|\mathbf{I})}
\end {equation}
where $\psi_u(X_i|\mathbf{I})$ is called unary potential. In our work, our computed $\mathbf{S}_{in}$ is treated as the unary potential:
\begin {equation}\label{eq:unarypotential}
\psi_u(X_i|\mathbf{I})=\mathbf{S}_{in}(i)
\end {equation}
And $\psi_p(X_i,X_j|\mathbf{I})$ is called pairwise potential. It accounts for the joint distribution of pixels $i$ and $j$. It allows us to explicitly model interactions between pixels, such as pixels with similar colour are
likely the same class. And $\psi_p$ is formulated as weighted sum of Gaussian kernels:
\begin {equation}\label{eq:pairpotential}
\psi_p(X_i,X_j|\mathbf{I})=\mu(X_i,X_j)k(\mathbf{f}_i,\mathbf{f}_j)
\end {equation}
where $\mu(X_i,X_j)$ is a simple label compatibility function, which is given by the Potts model $\mu(X_i,X_j)=[X_i\ne X_j]$. It
penalizes nearby similar pixels that are assigned different labels. $k(\mathbf{f}_i,\mathbf{f}_j)$ denotes Gaussian kernels with feature vectors $\mathbf{f}_i$ and $\mathbf{f}_j$ in an arbitrary feature space. Specifically, contrast-sensitive two-kernel potentials are formulated in our model:
\begin {equation}\label{eq:gaussiankernel}
\begin{aligned}
k(\mathbf{f}_i,\mathbf{f}_j)=&\underbrace{w^{(1)}\mathrm{exp}(-\frac{|\mathrm{p}_i-\mathrm{p}_j|^2}{2\theta_{\alpha}^2}-\frac{|\mathrm{I}_i-\mathrm{I}_j|^2}{2\theta_{\beta}^2})}_{\text{appearance kernel}}\\&+\underbrace{w^{(2)}\mathrm{exp}(-\frac{|\mathrm{p}_i-\mathrm{p}_j|^2}{2\theta_{\gamma}^2})}_{\text{smoothness kernel}}
\end{aligned}
\end {equation}
where $\mathrm{I}_i$ and $\mathrm{I}_j$ are color vectors, $\mathrm{p}_i$ and $\mathrm{p}_j$ are positions. $w^{(1)}$ and $w^{(2)}$ are linear combination weights. $\theta_{\alpha}$, $\theta_{\beta}$ and $\theta_{\gamma}$ are controlling parameters. The appearance kernel drives nearby pixels with similar color to be in the same class. The smoothness kernel removes small isolated regions.
In our implementation, we adopt ConvCRF \cite{teichmann2018convolutional} for inference. ConvCRF adds the assumptions of conditional independence fully-connected CRF, and reformulates the inference in terms of convolutions which are implemented highly efficiently on GPUs.

\section{Experimental Evaluation}
\label{sec:experiment}

\subsection{Implementation Details}
\label{ssec:impledetails}

Real-world copy-move forgery needs forgers to manually manipulate images and pasted regions. Therefore, the available copy-move forgery datasets are not sufficient for training an end-to-end deep matching network. Thus, we automatically generate a synthetic training set and a synthetic testing set from MS COCO 2014 training images and testing images respectively. For each image, we resize it to $512\times 512$, randomly select one annotated region under different transformations, and paste it to a random position of this image. All pasted regions randomly suffer four types of transformations, i.e., rotation changes in $\mathbb{U}(-60,60)$, scale changes in $\mathbb{U}(0.5,4)$, luminance changes in $\mathbb{U}(-32,32)$, and deformation changes in $\mathbb{U}(0.5,2)$ (decrease or increase the width of a tampered region). Following this strategy, we generate $120,000$ training images and $1,000$ testing images.

The self deep matching network is trained with a single spatial cross
entropy loss, and parameters in the basic feature extraction network are initialized using VGG16 \cite{simonyan2014very} which
is trained for image classification. Similarly, alternative formulations are initilized using corresponding classification networks. We conduct $16$-epoch training, and adopt the Adadelta optimizer \cite{zeiler2012adadelta}. The input images are randomly resized in the range of $[256\times 256,512\times 512]$. Limited by our GPU memory, the batch size is set to $6$ (a larger batch size may further improve our performance). As for the Proposal SuperGlue stage, no further training is needed. We directly adopt their trained DeepMask \cite{pinheiro2015learning}, SuperPoint \cite{detone2018superpoint}, SuperGlue \cite{sarlin2020superglue} models.

\subsection{Backbone Network Ablation Study}
\label{ssec:ablation} 

\begin{table*}[htp]
	\renewcommand{\arraystretch}{1.3}
	\caption{Step-by-step analyses on the synthetic testing set.}
	\label{table:combstep}
	\centering
	\footnotesize
	\setlength{\tabcolsep}{1.6mm}{
	\begin{tabular}{c | c c c c | c c c c | c c }
		\hline
		\multirow{2}{*}{Variant} & \multicolumn{4}{c|}{Protocol-All} & \multicolumn{4}{c|}{Protocol-Detected} & \multirow{2}{*}{$T$} & \multirow{2}{*}{Trainable params} \\\cline{2-9}
		& IoU & Precision & Recall & F1-score & IoU & Precision & Recall & F1-score & & \\
		\hline
		encoder-decoder & 0.4982 & 0.5930 & 0.7905 & 0.6328 & 0.4992 & 0.5942 & 0.7921 & 0.6341 & 48 & 7,772,209 \\
		encoder-decoder-skip & 0.5523 & 0.6523 & 0.7927 & 0.6835 & 0.5529 & 0.6530 & 0.7935 & 0.6841 & 48 & 14,985,265  \\
		encoder-decoder-skip-normfeas & 0.6822 & 0.7793 & 0.8332 & 0.7897 & 0.6822 & 0.7793 & 0.8332 & 0.7897 & 48 & 14,985,265\\
		SelfDM & 0.6959 & 0.7813 & 0.8506 & 0.7999 & 0.6959 & 0.7813 & 0.8506 & 0.7999 & 48 & 14,985,265 \\
		\hline
		SelfDM-16 & 0.6818 & 0.7703 & 0.8427 & 0.7891 & 0.6832 & 0.7719 & 0.8444 & 0.7907 & 16 & 14,851,633 \\
		SelfDM-32 & 0.6881 & 0.7899 & 0.8255 & 0.7927 & 0.6881 & 0.7899 & 0.8255 & 0.7927 & 32 & 14,918,449 \\
		SelfDM-48 & 0.6959 & 0.7813 & 0.8506 & 0.7999 & 0.6959 & 0.7813 & 0.8506 & 0.7999 & 48 & 14,985,265 \\
		SelfDM-64 & 0.6942 & 0.8038 & 0.8221 & 0.7970 & 0.6949 & 0.8046 & 0.8229 & 0.7978 & 64 & 15,052,081 \\
		\hline
		SelfDM-SA & 0.7233 & 0.8458 & 0.8227 & 0.8216 & 0.7240 & 0.8467 & 0.8235 & 0.8225 & 48 & 15,724,148 \\
		SelfDM-MSA & 0.7119 & 0.8466 & 0.8064 & 0.8096 & 0.7126 & 0.8475 & 0.8072 & 0.8104 & 48 & 17,940,797 \\
		SelfDM-SACA & 0.7129 & 0.8370 & 0.8204 & 0.8136 & 0.7129 & 0.8370 & 0.8204 & 0.8136 & 48 & 15,799,236 \\
		SelfDM-CASA & 0.7195 & 0.8472 & 0.8164 & 0.8182 & 0.7195 & 0.8472 & 0.8164 & 0.8182 & 48 & 15,799,236 \\
		\hline
	\end{tabular}}
\end{table*}

\begin{table*}[htp]
	\renewcommand{\arraystretch}{1.3}
	\caption{Feature extractor comparisons on the synthetic testing set.}
	\label{table:combmodel}
	\centering
	\footnotesize
	\setlength{\tabcolsep}{2.35mm}{
	\begin{tabular}{c | c c c c | c c c c | c }
		\hline
		\multirow{2}{*}{Variant} & \multicolumn{4}{c|}{Protocol-All} & \multicolumn{4}{c|}{Protocol-Detected} & \multirow{2}{*}{Trainable params} \\\cline{2-9}
		& IoU & Precision & Recall & F1-score & IoU & Precision & Recall & F1-score &  \\
		\hline
		SelfDM-SA & 0.7233 & 0.8458 & 0.8227 & 0.8216 & 0.7240 & 0.8467 & 0.8235 & 0.8225 & 15,724,148 \\
		SelfDM-SA-ResNet50 & 0.7372 & 0.7809 & 0.9191 & 0.8312 & 0.7372 & 0.7809 & 0.9191 & 0.8312 & 30,664,372 \\
		SelfDM-SA-ResNet101 & 0.7176 & 0.8246 & 0.8359 & 0.8059 & 0.7183 & 0.8254 & 0.8368 & 0.8067 & 49,656,500 \\
		SelfDM-SA-MobileNetV2 & 0.7126 & 0.7957 & 0.8584 & 0.8118 & 0.7126 & 0.7957 & 0.8584 & 0.8118 & 4,557,012 \\
		SelfDM-SA-MobileNetV3 & 0.7512 & 0.8575 & 0.8467 & 0.8412 & 0.7512 & 0.8575 & 0.8467 & 0.8412 & 4,092,268 \\
		SelfDM-SA-ShuffleNetV2 & 0.6676 & 0.7692 & 0.8137 & 0.7745 & 0.6676 & 0.7692 & 0.8137 & 0.7745 & 2,920,602  \\
		\hline
	\end{tabular}}
\end{table*}

Our backbone self deep matching network incorporates encoder-decoder architecture with atrous convolution, skip matching, feature normalization, correlation normalization and spatial attention. In TABLE~\ref{table:combstep}, step-by-step analyses are provided on the synthetic testing set. We compute the pixel-level IoU, precision, recall and F1-score for each image, and caculate their average scores. Two protocols are adopted: ``Protocol-All" means we compute the average scores of all evaluated images, and ``Protocol-Detected" only computes average scores of detected images.

It shows that each component of our backbone network plays an important role in improving its localization performance. Specifically, ``encoder-decoder" denotes a pure architecture with a feature extractor and a decoder, there is no skips, normalization and attention; In ``encoder-decoder-skip", we add skip matching; In ``encoder-decoder-skip-normfeas",  input feature maps before correlation computation are normalized; In ``SelfDM", correlation maps are followed by ReLU and L2-normalization. Besides, we make a discussion on the selection of $T$ in Eq.~(\ref{eq:sortpool}). We find that it can even achieve comparable performance when $T=16$. When we set $T=48$, it can get higher scores. There is no further improvement with $T=64$. More importantly, we test four types of attention-based self-correlation formulations, i.e., ``SelfDM-SA" with spatial attention, ``SelfDM-MSA" with multi-head spatial attention \cite{vaswani2017attention}, ``SelfDM-SACA" with spatial attention before channel attention and ``SelfDM-CASA" with spatial attention after channel attention \cite{hu2018squeeze}. Although, ``SelfDM-MSA", ``SelfDM-SACA" and ``SelfDM-CASA" have more parameters, their performance is barely satisfactory. We guess the main reasons are that: (1) single spatial attention can already reinforce correlated regions, while multiple spatial attention with redundant information may mislead our model; (2) SE blocks \cite{hu2018squeeze} (channel attention) improve the classification performance by densely adding them into convolutional blocks, while we only add them into self-correlation which is not sufficient enough. After comprehensive comparison, we select the version with spatial attention, i.e., ``SelfDM-SA".

In section \ref{sssec:FEAC}, alternative formulations are discussed. Two deeper networks (ResNet50, ResNet101) and three light-weight networks (MobileNetV2, MobileNetV3, ShuffleNetV2) are consturcted. We find that ``SelfDM-SA-ResNet50" can slightly improve the performance of our backbone network, while the performance of deeper ``SelfDM-SA-ResNet101" is even worse. It shows that high-level features with richer semantic information are not as important as discriminative features with rich spatial information, in a deep matching task. Furthermore, light-weight networks are compared. The performance of ``SelfDM-SA-MobileNetV3" is even better. So we finally select the default ``SelfDM-SA" with VGG, ``SelfDM-SA-ResNet50" and ``SelfDM-SA-MobileNetV3" for further comparison in the next section.

\subsection{Comparison with State-of-the-art Methods}
\label{ssec:csms}

\begin{table*}[hbtp]
	\renewcommand{\arraystretch}{1.3}
	\caption{Comparisons with the state-of-the-art methods on the synthetic testing set.}
	\label{table:comparecomb}
	\centering
	\footnotesize
	\setlength{\tabcolsep}{3.2mm}{
		\begin{tabular}{c | c c c c | c c c c }
			\hline
			\multirow{2}{*}{Methods} & \multicolumn{4}{c|}{Protocol-All} & \multicolumn{4}{c}{Protocol-Detected} \\\cline{2-9}
			& IoU & Precision & Recall & F1-score & IoU & Precision & Recall & F1-score \\
			\hline
			LiJ \cite{li2015segmentation} & 0.3188 & 0.3620 & 0.3723 & 0.3597 & 0.6826 & 0.7752 & 0.7972 & 0.7702 \\
			Cozzolino \cite{cozzolino2015efficient} & 0.2377 & 0.3105 & 0.2584 & 0.2728 & 0.6911 & 0.9027 & 0.7513 & 0.7931 \\
			BusterNet \cite{wu2018busternet} & 0.3349 & 0.5814 & 0.3764 & 0.4213 & 0.4412 & 0.7660 & 0.4959 & 0.5550 \\
			\hline
			SelfDM-SA & 0.7233 & 0.8458 & 0.8227 & 0.8216 & 0.7240 & 0.8467 & 0.8235 & 0.8225 \\
			SelfDM-SA+PS & 0.7198 & 0.8445 & 0.8232 & 0.8186 & 0.7205 & 0.8453 & 0.8239 & 0.8194 \\
			SelfDM-SA+PS+CRF & 0.7403 & 0.8785 & 0.8176 & 0.8308 & 0.7433 & 0.8820 & 0.8209 & 0.8342 \\
			\textit{SelfDM-SA+CRF}$\star$ & \textit{0.7317} & \textit{0.8921} & \textit{0.7958} & \textit{0.8252} & \textit{0.7376} & \textit{0.8993} & \textit{0.8024} & \textit{0.8309} \\
			\hline
			SelfDM-SA-ResNet50 & 0.7372 & 0.7809 & 0.9191 & 0.8312 & 0.7372 & 0.7809 & 0.9191 & 0.8312 \\		
			SelfDM-SA-ResNet50+PS & 0.7247 & 0.7742 & 0.9112 & 0.8232 & 0.7247 & 0.7742 & 0.9112 & 0.8232 \\
			SelfDM-SA-ResNet50+PS+CRF & 0.7438 & 0.8066 & 0.8986 & 0.8358 & 0.7438 & 0.8066 & 0.8986 & 0.8358 \\
			\hline
			SelfDM-SA-MobileNetV3 & 0.7512 & 0.8575 & 0.8467 & 0.8412 & 0.7512 & 0.8575 & 0.8467 & 0.8412 \\
			SelfDM-SA-MobileNetV3+PS & 0.7364 & 0.8488 & 0.8391 & 0.8302 & 0.7364 & 0.8488 & 0.8391 & 0.8302 \\
			SelfDM-SA-MobileNetV3+PS+CRF & 0.7531 & 0.8848 & 0.8281 & 0.8394 & 0.7538 & 0.8856 & 0.8289 & 0.8403 \\
			\hline
	\end{tabular}}
\end{table*}

We adopt four datasets for comprehensive comparisons: our synthetic testing set, CoMoFoD dataset \cite{tralic2013comofod}, CASIA CMFD dataset \cite{wu2018busternet}, and MICC-F220 dataset \cite{amerini2011sift}.

\begin{figure}[htbp]
	\centering
	\includegraphics[width=0.99999\columnwidth]{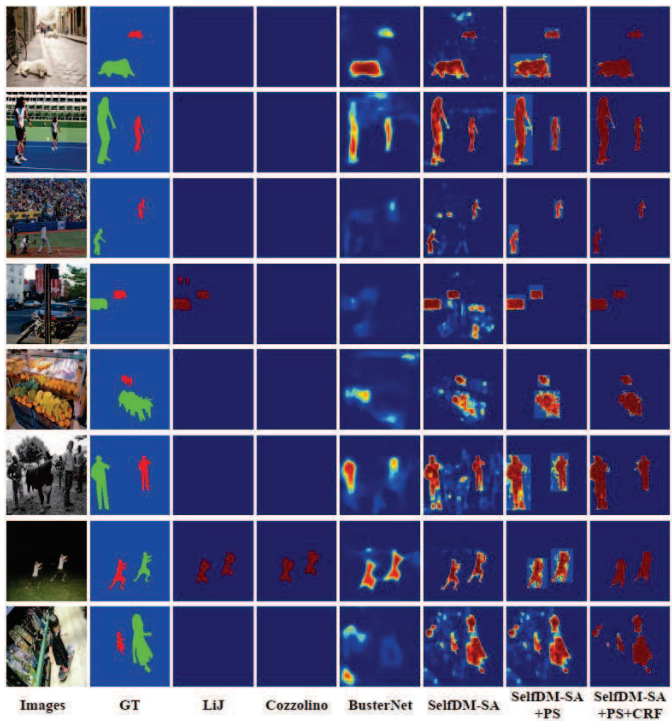}
	\caption{Example copy-move forgery detection results on the synthetic testing set. In the ``GT" column, red regions indicate forged regions and green regions are original regions.}
	\label{Figure:cocovis}
\end{figure}

Pasted regions in our synthetic testing set have gone through multiple changes with greater extent. Besides, there is only one pair of similar regions in each image, and there is no obvious ``disturbance" (similar but genuine regions) for the most cases. So it can indicate the capability and robustness of algorithms to detect appearance similar regions under different transformations. We select a representative keypoint-based method (LiJ \cite{li2015segmentation}), an advanced block-based method (Cozzolino \cite{cozzolino2015efficient}), and an end-to-end deep learning method (BusterNet \cite{wu2018busternet}) for comparison. Scores in TABLE~\ref{table:comparecomb} are generated by their codes provided by the authors. It clearly shows that classical methods (``LiJ" and ``Cozzolino") are not robust enough against different transformations, while their detected regions mostly are accurate (comparable scores in ``Protocol-Detected"). Our backbone network has strong ability to detect similar regions. Since there are few ``disturbances", the ability of Proposal SuperGlue to remove false-alarmed regions is not obvious in this dataset. However, it still can be seen that precisions are increased and overall scores are higher. Specifically, ``SelfDM-SA+PS" indicates the two-stage version without ConvCRF optimization, ``SelfDM-SA+PS+CRF" adds ConvCRF, and ``SelfDM-SA+CRF$\star$" directly conducts ConvCRF on the backbone network which is used to demonstrate the effectiveness of Proposal SuperGlue. Furthermore, visual comparions are provided in Fig.~\ref{Figure:cocovis}. In rows 1 and 2, we provide images with obvious scale changes, SelfDM-SA can already achieve satisfied performance. In the column of ``SelfDM-SA+PS", proposals (light blue rectangular regions with lower scores) can enclose suspected regions, and further SuperGlue with SuperPoint can be conducted. With the help of ConvCRF, the score distribution can be optimized. BusterNet only detects their approximate locations without accurate boundaries. And it is difficult for classical methods to detect regions under severe transformations like scale or rotation changes. In rows 3 to 6, we provide four challengable cases. Our backbone network detects some false-alarmed regions or only detects partial regions, while proposal SuperClue can enclose those regions and clearly optimize the detected regions. In the last two rows, we also provide two failure cases. SelfDM-SA can already generate an accurate result while Proposal SuperGlue causes some false-alarmed regions around boundaries in row 7. Or no meaningful proposals are obtained in row 8.

\begin{figure}[htbp]
	\centering
	\includegraphics[width=0.99999\columnwidth]{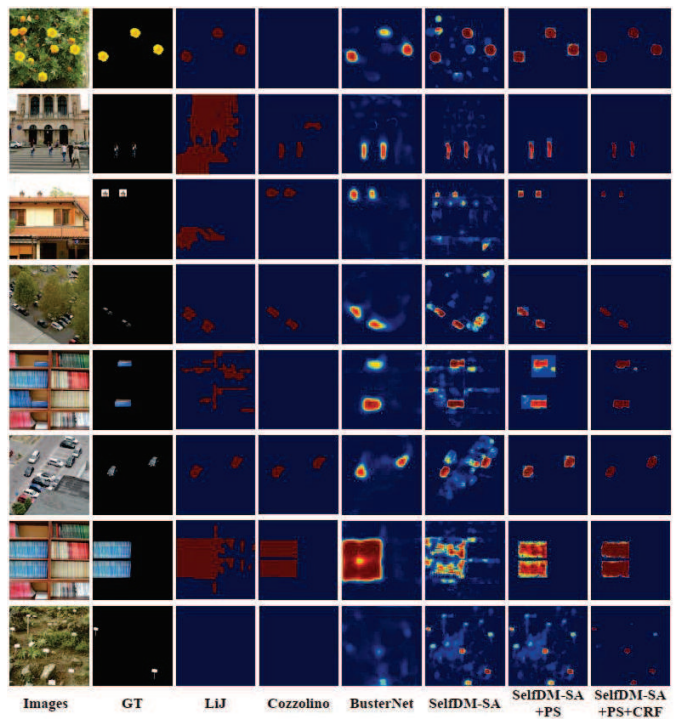}
	\caption{Example copy-move forgery detection results on CoMoFoD.}
	\label{Figure:comovis}
\end{figure}

\begin{figure}[htp]
	\begin{minipage}[b]{0.8\linewidth}
		\centering
		\centerline{\includegraphics[width=5cm]{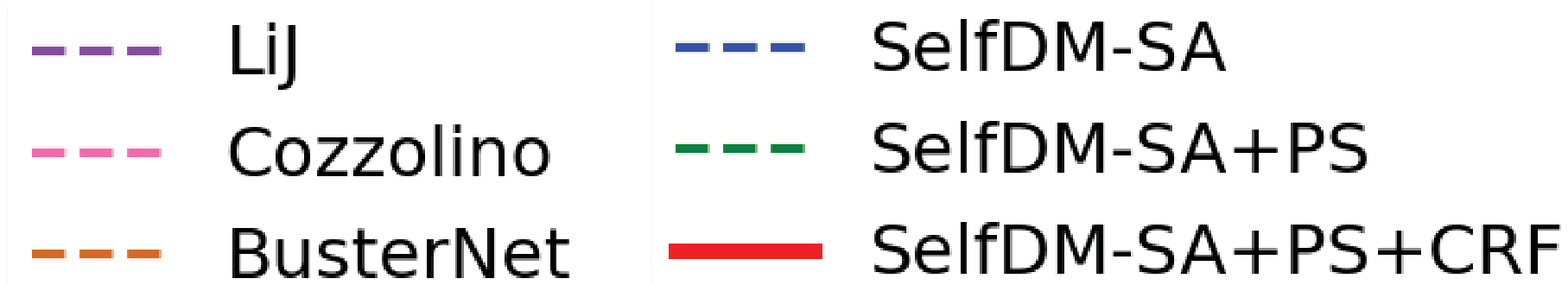}}
	\end{minipage}
	\begin{minipage}[b]{0.48\linewidth}
		\centering
		\centerline{\includegraphics[width=5cm]{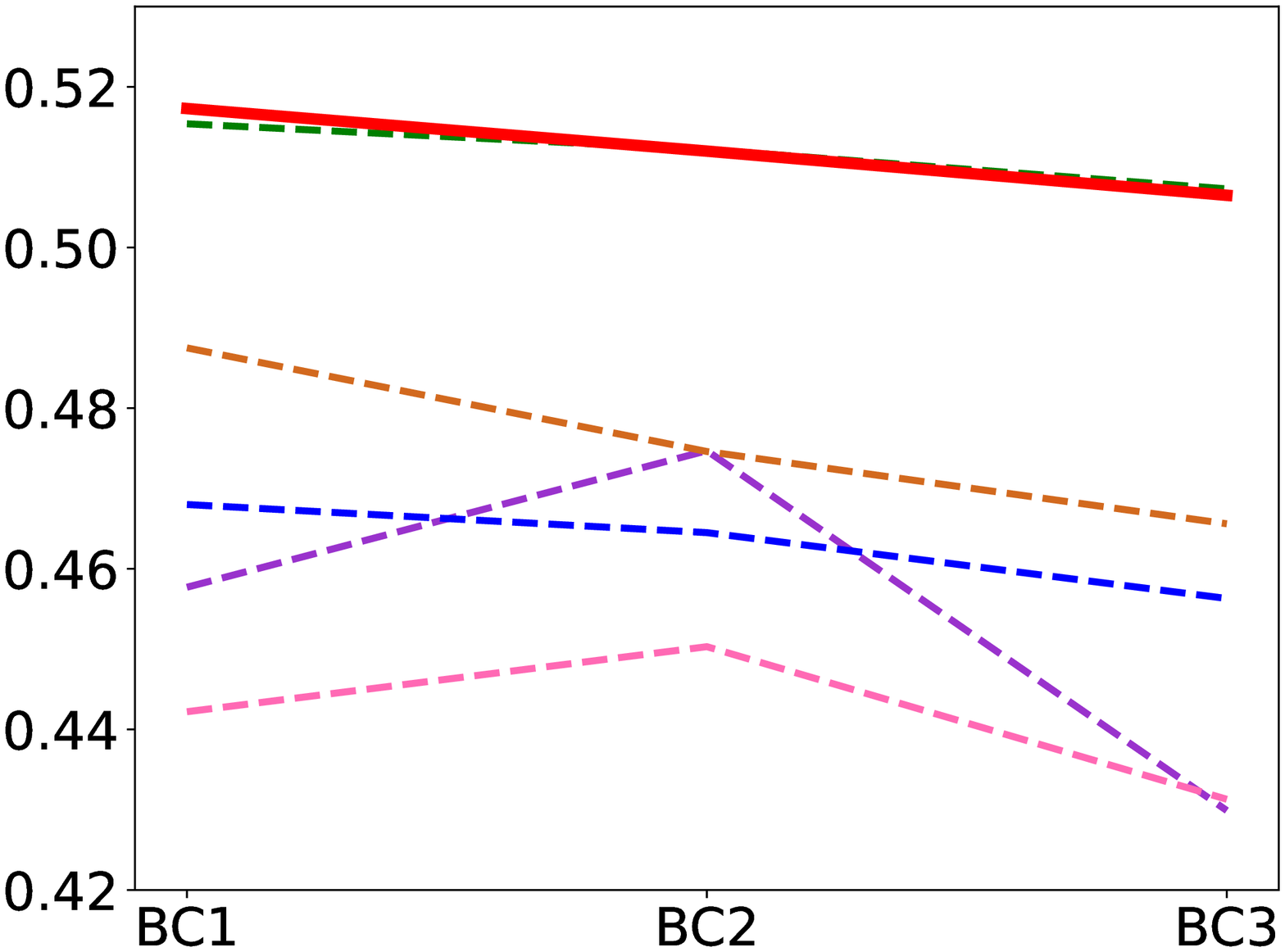}}
		\centerline{\footnotesize{Brightness change (BC)}}
	\end{minipage}
	\hfill
	\begin{minipage}[b]{0.48\linewidth}
		\centering
		\centerline{\includegraphics[width=5cm]{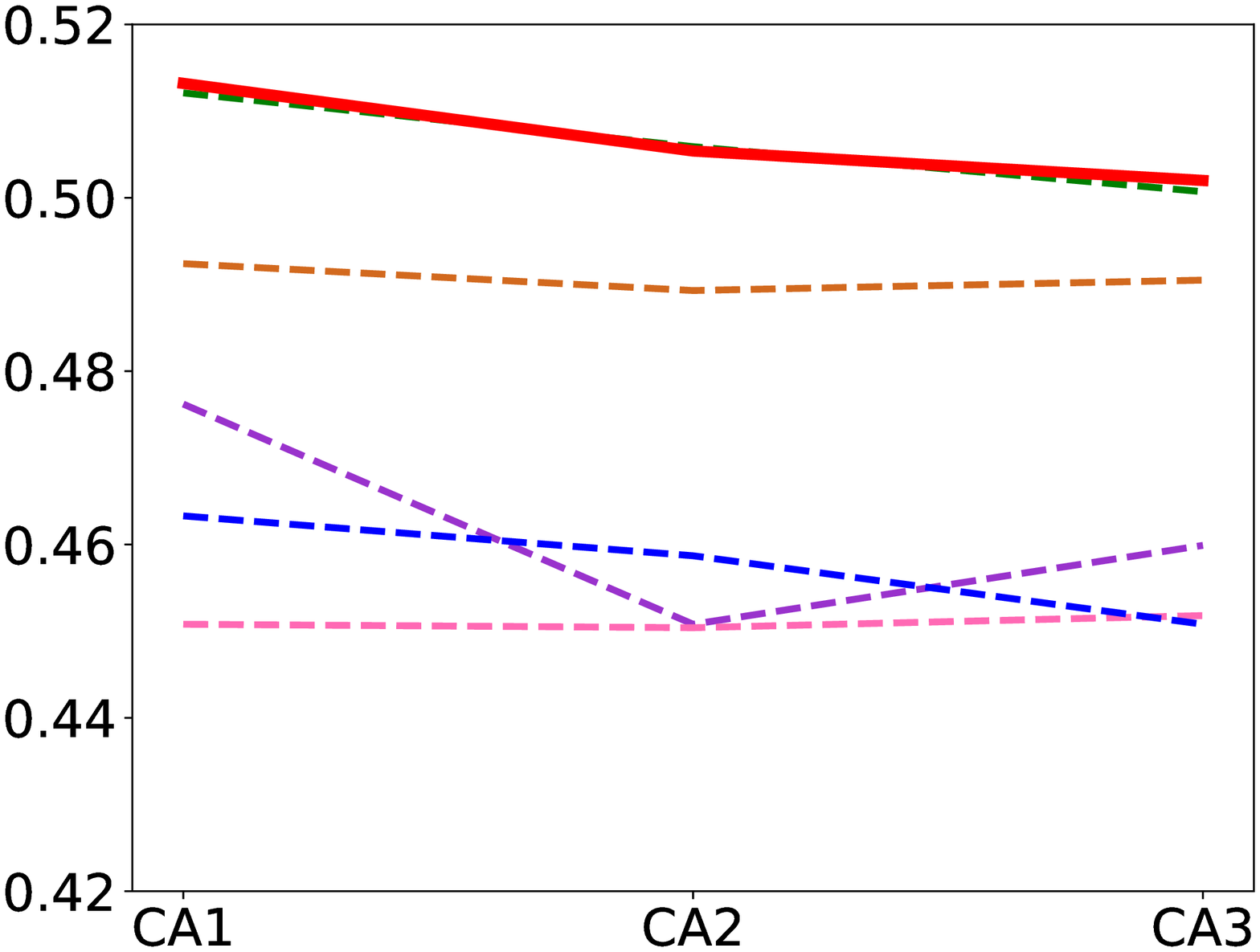}}
		\centerline{\footnotesize{Contrast adjustments (CA)}}
	\end{minipage}
	\vfill
	\begin{minipage}[b]{0.48\linewidth}
		\centering
		\centerline{\includegraphics[width=5cm]{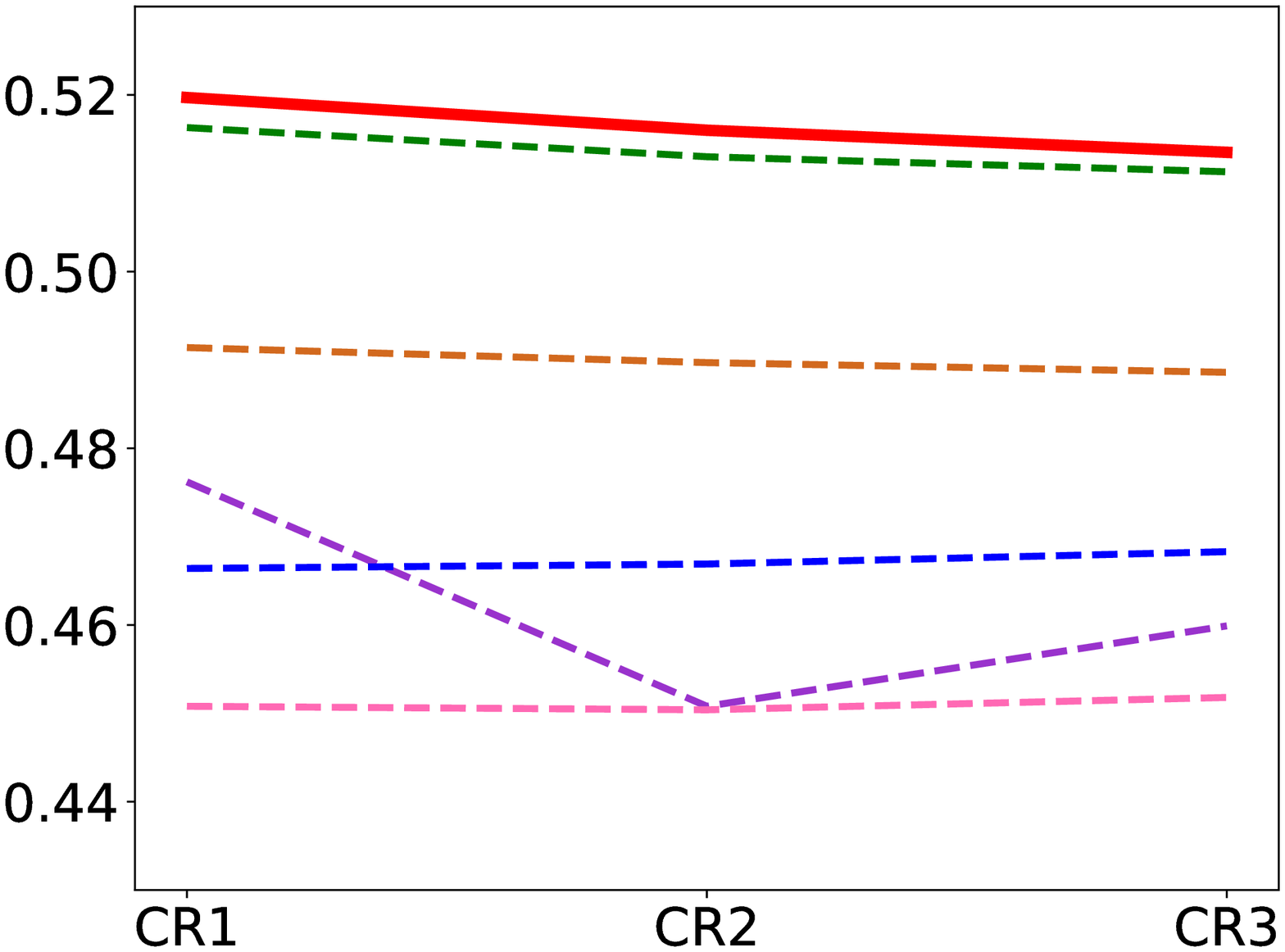}}
		\centerline{\footnotesize{Color reduction (CR)}}
	\end{minipage}
	\hfill
	\begin{minipage}[b]{0.48\linewidth}
		\centering
		\centerline{\includegraphics[width=5cm]{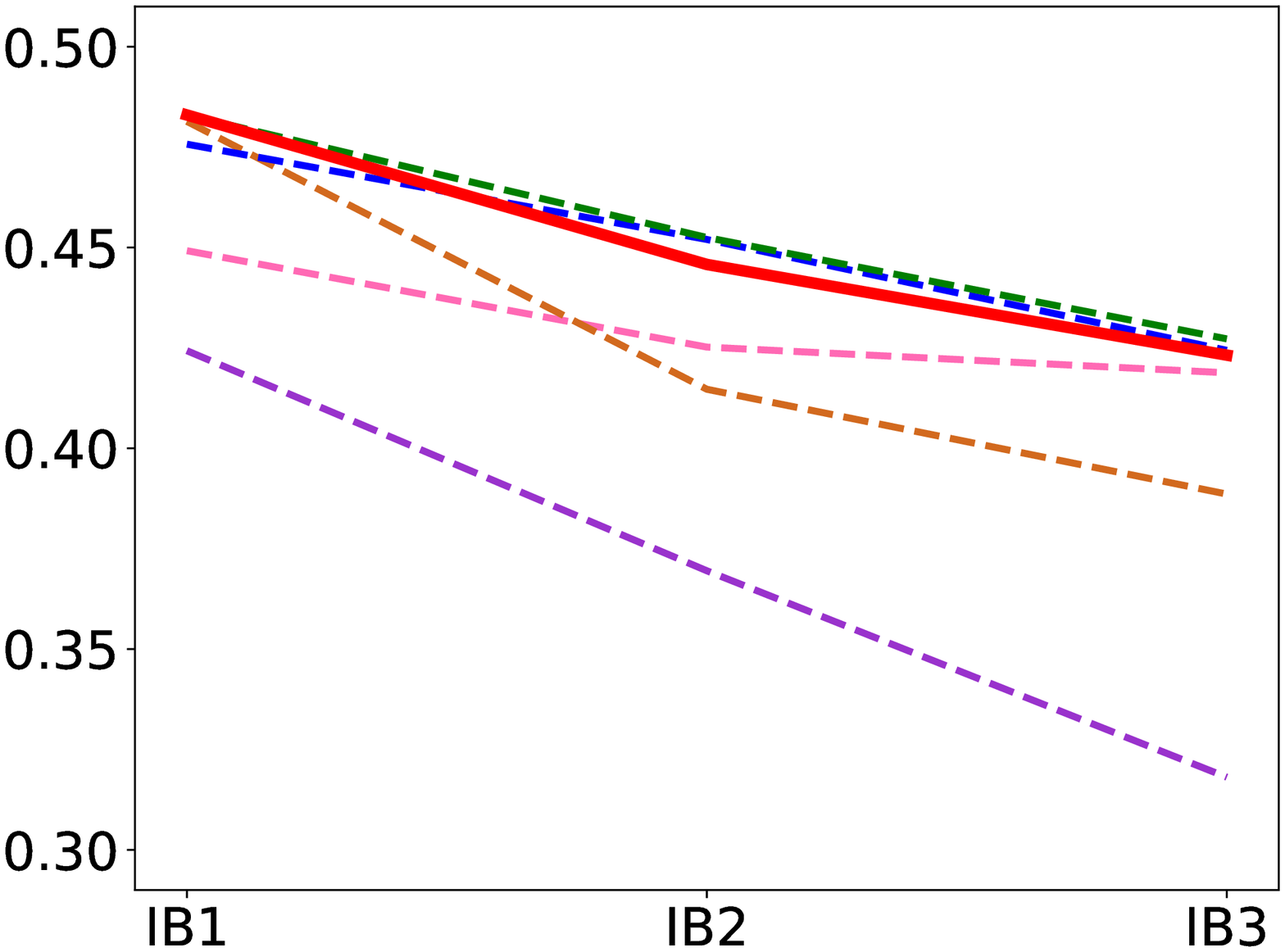}}
		\centerline{\footnotesize{ Image blurring (IB)}}
	\end{minipage}
	\vfill
	\begin{minipage}[b]{0.48\linewidth}
		\centering
		\centerline{\includegraphics[width=5cm]{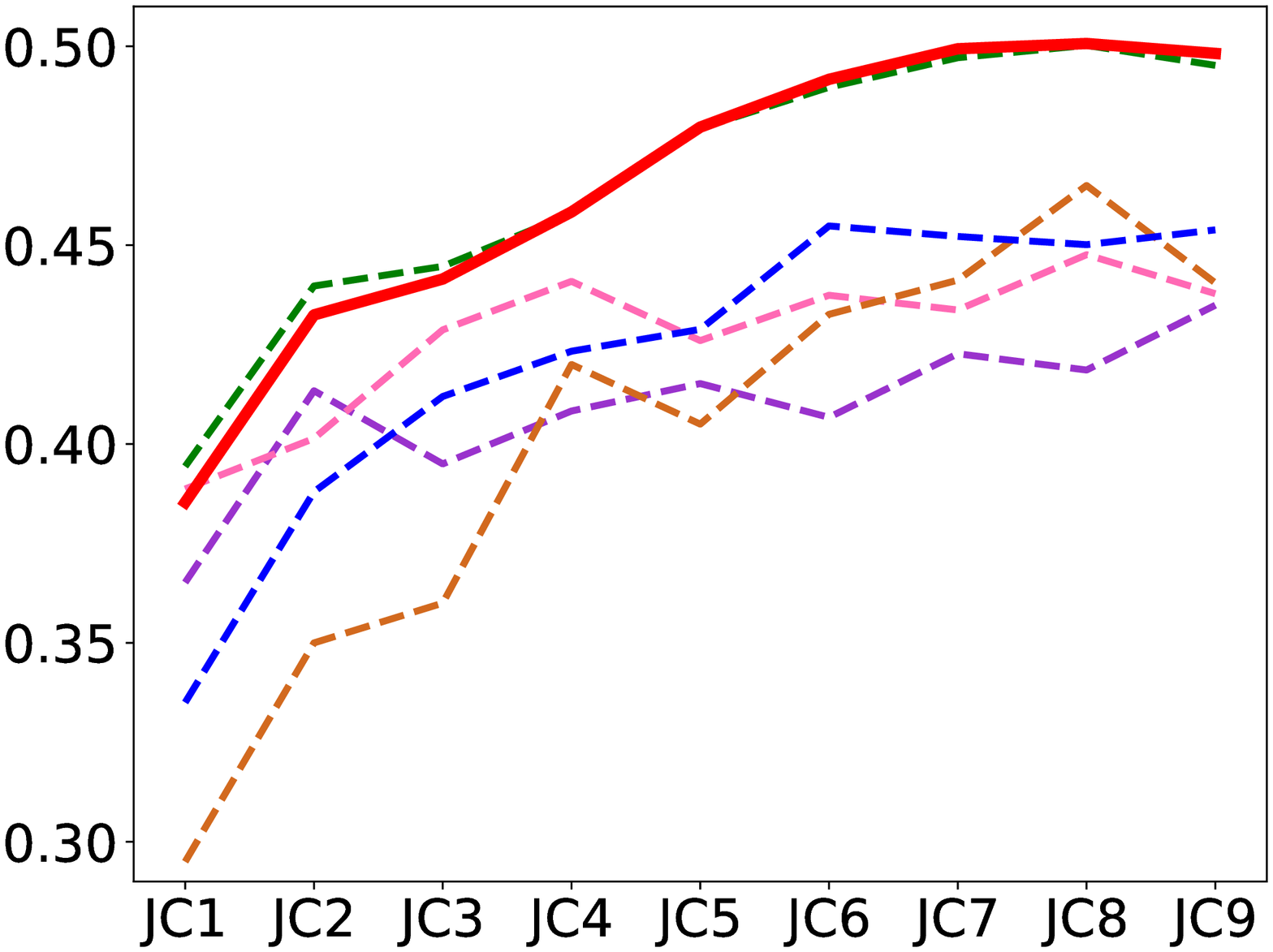}}
		\centerline{\footnotesize{JPEG compression (JC)}}
	\end{minipage}
	\hfill
	\begin{minipage}[b]{0.48\linewidth}
		\centering
		\centerline{\includegraphics[width=5cm]{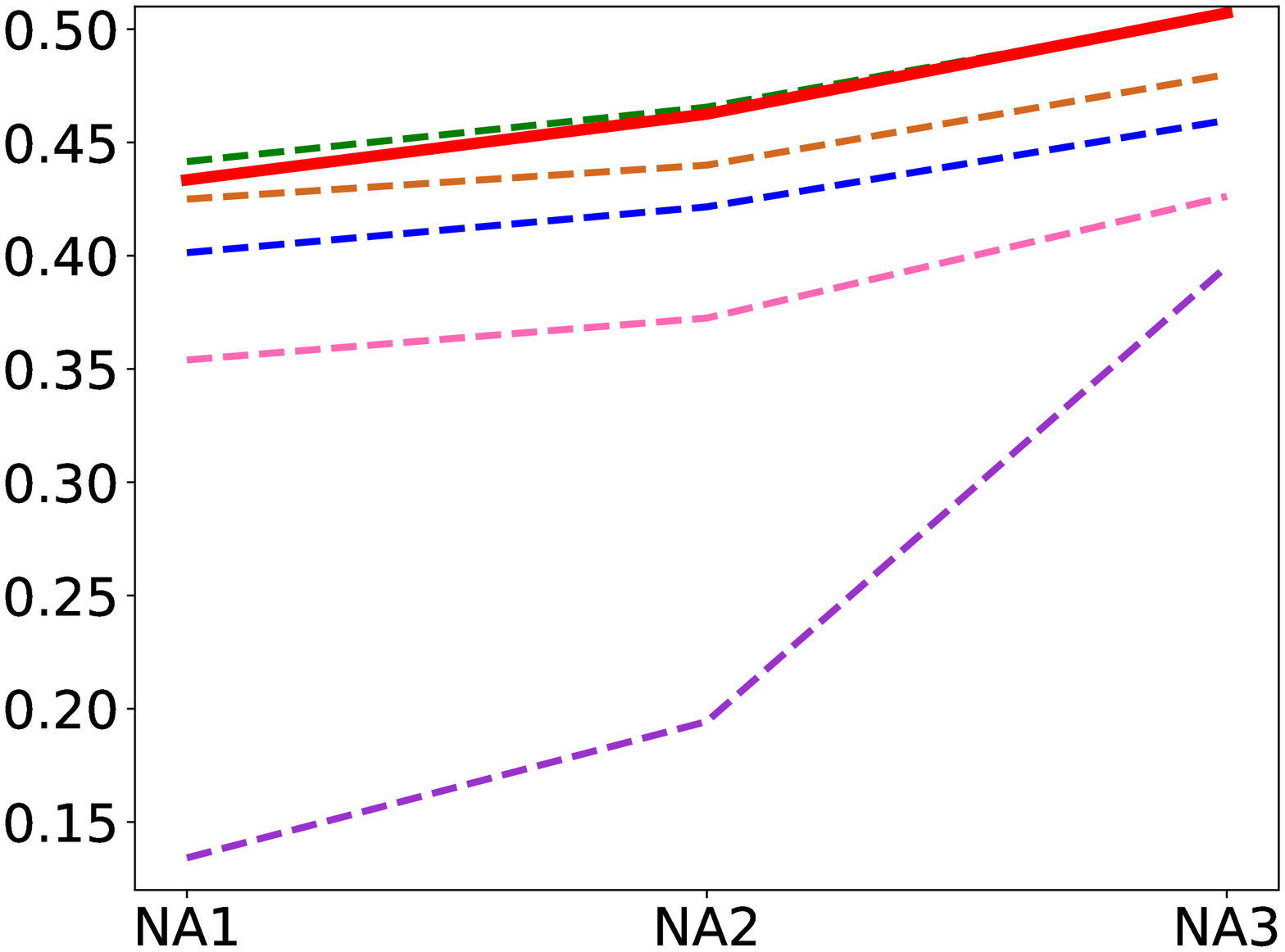}}
		\centerline{\footnotesize{Noise adding (NA)}}
	\end{minipage}
	\caption{Pixel-level F1 scores (y-axis) on	CoMoFoD under attacks (x-axis).}
	\label{Figure:f1scorescomofodattacks}
\end{figure}

The CoMoFoD dataset consists of $200$ copy-move forged images with resolution $512\times 512$. Besides the version with no postprocessing, these images are processed under $6$ kinds of postprocessing respectively, namely brightness change (BC), contrast adjustments (CA), color reduction (CR), image
blurring (IB), JPEG compression (JC), and noise adding (NA). In TABLE~\ref{table:CoMoFoDnoattack}, two measure protocols are used. ``Correctly Detected Average" only computes the average score of correctly detected images. An image is referred as correctly detected if its pixel-level F1-score is higher
than $0.5$. ``Overall Average" computes their average scores of all images. We find that although our backbone network can achieve excellent performance on synthetic testing images, there is no obvious advantage on CoMoFoD. The main reason is that pasted regions in CoMoFod have gone through only slight changes, compared methods can already achieve good performance. Besides, limited by the training set, the majority of synthetic images only have a pair of similar objects. In another word, there are few of disturbances with similar appearance. Our backbone network has strong ability to detect similar objects under severe transformations, so that there are inevitably some false-alarmed regions (rows 1, 3, 4, 5, 6, 8 in Fig.~\ref{Figure:comovis}), which affect the scores on CoMoFoD. With the help of Proposal SuperGlue, it can be clearly seen that we can remove some false-alarmed regions and complement miss-detected regions (Fig.~\ref{Figure:comovis}). Thus, the performance can be obviously improved. Besides, the robustness against different postprocessing is evaluated and shown in Fig.~\ref{Figure:f1scorescomofodattacks}. Our two-stage version, i.e., SelfDM-SA+PS+CRF, can achieve consistently higher scores under different attacks, which also demonstrate the high robustness of our method. As for the alternative networks, i.e., SelfDM-SA-ResNet50 and SelfDM-SA-MobileNetV3, our Proposal SuperGlue can also notably improve their performance. Especially, the precision scores are improved.

CASIA CMFD dataset is selected from CASIA TIDEv2.0 dataset by Wu et al. \cite{wu2018busternet}. There are $1313$ CMFD samples and their authentic counterparts. They provide $256\times 256$ images and masks as a HDF dataset. In our experiments, both pixel-level and image-level scores are computed. Pixel-level scores are the overall average scores of all CMFD samples (the same as ``Overall Average" in TABLE~\ref{table:CoMoFoDnoattack}). Our backbone network based on VGG, i.e., SelfDM-SA, can already achieve higher scores, and Proposal SuperGlue with ConvCRF can further improve its performance. Dramatically, the MobileNetV3 version can achieve the best performance. It further demonstrates that the discriminative capability of features are more important in the deep matching task, and it is different from image understanding tasks (e.g. image classification, object detection, semantic segmentation) in which high-level features with more semantic informantion play a more important role.

\begin{table*}[htp]
	\renewcommand{\arraystretch}{1.3}
	\caption{Comparisons on CoMoFoD dataset with no attack.}
	\label{table:CoMoFoDnoattack}
	\centering
	\footnotesize
	\setlength{\tabcolsep}{3.55mm}{
	\begin{tabular}{c | c c c c | c c c }
		\hline
		\multirow{2}{*}{Method} & \multicolumn{4}{c|}{Correctly Detected Average} & \multicolumn{3}{c}{Overall Average} \\\cline{2-8}
		& Detected Rate & Precision & Recall & F1-score & Precision & Recall & F1-score \\
		\hline
		Ryu2010\cite{ryu2010detection} & 0.450 & 0.9627 & 0.6984 & 0.7993 & 0.4578 & 0.3435 & 0.3737 \\
		LiJ \cite{li2015segmentation} & 0.510 & 0.8042 & 0.9586 & 0.8616 & 0.4247 & 0.6633 & 0.4644 \\
		Cozzolino \cite{cozzolino2015efficient} & 0.505 & 0.8132 & 0.9384 & 0.8591 & 0.4174 & 0.5042 & 0.4440 \\
		Wu2018 \cite{wu2018image} & 0.265 & 0.6111 & 0.7148 & 0.6313 & 0.3629 & 0.4041 & 0.3113 \\
		BusterNet \cite{wu2018busternet} & 0.585 & 0.8352 & 0.7875 & 0.8009 & 0.5734 & 0.4939 & 0.4926\\
		\hline
		SelfDM-SA & 0.475 & 0.7086 & 0.8210 & 0.7350 & 0.5722 & 0.5216 & 0.4660 \\
		SelfDM-SA+PS & 0.575 & 0.7895 & 0.8087 & 0.7641 & 0.6086 & 0.5624 & 0.5151 \\
		SelfDM-SA+PS+CRF & 0.545 & 0.8139 & 0.8282 & 0.7943 & 0.6375 & 0.5444 & 0.5172 \\
		\hline
		SelfDM-SA-ResNet50 & 0.520   & 0.7518 & 0.7394 & 0.7208 & 0.5340 & 0.5467 & 0.4753 \\
		SelfDM-SA-ResNet50+PS & 0.545 & 0.8439 & 0.7668 & 0.7786 & 0.5910 & 0.5631 & 0.5108 \\
		SelfDM-SA-ResNet50+PS+CRF & 0.555 & 0.8728 & 0.7432 & 0.7773 & 0.6231 & 0.5342 & 0.5088 \\
		\hline
		SelfDM-SA-MobileNetV3 & 0.465 & 0.6526 & 0.8468 & 0.7096 & 0.4833 & 0.5198 & 0.4299 \\
		SelfDM-SA-MobileNetV3+PS & 0.530 & 0.7488 & 0.8137 & 0.7438 & 0.5170 & 0.5260 & 0.4645 \\
		SelfDM-SA-MobileNetV3+PS+CRF & 0.505 & 0.7885 & 0.8202 & 0.7742 & 0.5600 & 0.5049 & 0.4695 \\
		\hline
	\end{tabular}}
\end{table*}

\begin{table*}[htp]
	\renewcommand{\arraystretch}{1.3}
	\caption{Performance analysis on CASIA CMFD dataset.}
	\label{table:casia}
	\centering
	\footnotesize
	\setlength{\tabcolsep}{5.3mm}{
	\begin{tabular}{c | c c c | c c c}
		\hline
		\multirow{2}{*}{Method} & \multicolumn{3}{c|}{Pixel Level} & \multicolumn{3}{c}{Image Level} \\\cline{2-7}
		 & Precision & Recall & F1-score & Precision & Recall & F1-score \\
		\hline
		Ryu2010 \cite{ryu2010detection} & 0.2271 & 0.1336 & 0.1640 & 0.9701 & 0.2447 & 0.3908 \\
		Christlein \cite{christlein2012evaluation} & 0.3709 &  0.0014 & 0.0023 & 0.6849 & 0.6782 & 0.6815 \\
		Cozzolino \cite{cozzolino2015efficient} & 0.2492 & 0.2681 & 0.2543 & 0.9951 & 0.3061 & 0.4682  \\
		Wu2018 \cite{wu2018image}  & 0.2397 & 0.1379 & 0.1464 & 0.6637 & 0.7359 & 0.6980 \\
		BusterNet-simi \cite{wu2018busternet}  & 0.4723 & 0.4844 &  0.4372 & 0.7153 & 0.8073 & 0.7585 \\
		BusterNet \cite{wu2018busternet}  & 0.5571 & 0.4383 &  0.4556 & 0.7822 & 0.7389 & 0.7598 \\
		\hline
		SelfDM-SA & 0.6551 & 0.4353 & 0.4635 & 0.7707 & 0.7807 & 0.7757 \\
		SelfDM-SA+PS & 0.6485 & 0.4531 & 0.4709 & 0.7860 & 0.7609 & 0.7732 \\
		SelfDM-SA+PS+CRF & 0.6494 & 0.4520 & 0.4782 & 0.7860 & 0.7609 & 0.7732 \\
		\hline
		SelfDM-SA-ResNet50 & 0.5358 & 0.4500 & 0.4356 & 0.6038 & 0.9238 & 0.7303 \\
		SelfDM-SA-ResNet50+PS & 0.5359 & 0.4676 & 0.4464 & 0.6164 & 0.9018 & 0.7323 \\
		SelfDM-SA-ResNet50+PS+CRF & 0.5729 & 0.4679 & 0.4595 & 0.6164 & 0.9018 & 0.7323 \\
		\hline
		SelfDM-SA-MobileNetV3 & 0.6248 & 0.4778 & 0.4843 & 0.6914 & 0.8294 & 0.7542\\
		SelfDM-SA-MobileNetV3+PS & 0.6272 & 0.4856 & 0.4891 & 0.7040 & 0.8096 & 0.7531 \\
		SelfDM-SA-MobileNetV3+PS+CRF & 0.6362 & 0.4752 & 0.4918 & 0.7040 & 0.8096 & 0.7531 \\
		\hline
	\end{tabular}}
\end{table*}

\begin{table}[!t]
	\renewcommand{\arraystretch}{1.3}
	\caption{Image-level performance on MICC-F220 dataset.}
	\label{table:miccf220}
	\centering
	\footnotesize
	\begin{tabular}{c | c c c}
		\hline
		Method & TPR & FPR & F1-score \\
		\hline
		Cozzolino \cite{cozzolino2015efficient} & 0.8455 & 0.1727 & 0.8378 \\
		LiJ \cite{li2015segmentation} & 0.7091 & 0.1727 & 0.7536 \\
		GoDeep \cite{silva2015going} & 0.4545 & 0.4182 & 0.4854 \\
		Zandi \cite{zandi2016iterative} & 0.7818 & 0.4818 & 0.6908 \\
		BusterNet \cite{wu2018busternet} & 0.4909 & 0.2000 & 0.5806 \\
		SelfDM-SA & 0.9273 & 0.2545 & 0.8500 \\
		SelfDM-SA+PS+CRF & 0.9182 & 0.2272 & 0.8559 \\
		SelfDM-SA-ResNet50 & 0.9727 & 0.6909 & 0.7304 \\
		SelfDM-SA-ResNet50+PS+CRF & 0.9545 & 0.6454 & 0.7343 \\
		SelfDM-SA-MobileNetV3 & 0.9636 & 0.4182 & 0.8092 \\
		SelfDM-SA-MobileNetV3+PS+CRF & 0.9545 & 0.3909 & 0.8140 \\
		\hline
	\end{tabular}
\end{table}

MICC-F220 is composed by $220$ images: $110$ tampered images and $110$ originals. There is no ground truth, and we evaluate the image-level performance. All the methods are evaluated by True Positive Rate (TPR), False Positive Rate (FPR) and corresponding F1-score, which are computed as: $TPR=TP/(TP+FN)$, $FPR=FP/(TN+FP)$, $F_1=2TP/(2TP+FP+FN)$, where $TP$ denotes true positive, $TN$ denotes true negtive, $FN$ denotes false negative and $FP$ denotes false positive. SelfDM-SA and the corresponding Proposal SuperGlue version can achieve better performance than many other state-of-the-art methods. Especially, we can achieve higher TPRs. However, the alternative formulations have higher FPRs, FPRs of the ResNet50 version are even greater than $0.6$. Considering all the experimental results on four datasets, we do not recommend the use of deeper networks for feature extraction, because they have more parameters, low efficiency and high false-alarmed rates. The VGG version is more stable and robust. The MobileNetV3 version has less parameters to learn and can achieve comparable performance on different datasets. Even so, we find that our two-stage framework can be applied to backbone networks with different feature extractors to achieve better performance.

\section{Conclusion}
\label{sec:conclusion}

In this paper, we propose a two-stage framework for deep learning based copy-move forgery detection. Our two-stage framework integrates self deep matching and keypoint matching by obtaining highly suspected proposals. The first stage is a backbone network which adopts atrous convolution, skip matching, and spatial attention. In the second stage, our Proposal SuperGlue is proposed to remove false alarms and complement incomplete regions. Specifically, we build a proposal selection module to enclose suspected regions, and conduct pairwise matching based on SuperPoint and SuperGlue. Integrated score map generation and refinement methods are proposed to obtain final results. Our two-stage framework can achieve consistently better performance on different public datasets. The two-stage framework relies on the performance of the backbone network. In the future, our two-stage framework can be further improved by designing a more powerful backbone network.


%





\ifCLASSOPTIONcaptionsoff
  \newpage
\fi



\bibliographystyle{IEEEtran}
\bibliography{mybibfile}
%



%




\end{document}